\documentclass[runningheads]{llncs}
\usepackage[utf8]{inputenc}
\usepackage[T1]{fontenc}
\usepackage[english]{babel}
\usepackage{textcomp}  
\usepackage[pdftex]{graphicx}
\usepackage{amsmath}
\usepackage{amssymb}
\usepackage{bm}
\usepackage{nicefrac}
\usepackage{microtype}
\usepackage{csquotes}
\usepackage[table]{xcolor} 
\usepackage{booktabs}
\usepackage{hyperref}
\hypersetup{pdfauthor={Sebastian P{\"{o}}lsterl, Ignacio Sarasua, Benjam{\'i}n Guti{\'e}rrez-Becker, Christian Wachinger}, pdftitle={A Wide and Deep Neural Network for Survival Analysis from Anatomical Shape and Tabular Clinical Data}}

\renewcommand{\bbbr}{\ensuremath\mathbb{R}}

\newcommand{\set}[1]{\ensuremath\mathcal{#1}}

\DeclareMathOperator*{\argmin}{arg\,min}
\begin{document} %
\title{A Wide and Deep Neural Network for Survival Analysis from Anatomical Shape and Tabular
Clinical Data}
\titlerunning{Wide and Deep Neural Network for Survival Analysis} %
\author{Sebastian P{\"{o}}lsterl \and Ignacio Sarasua \and Benjam{\'i}n Guti{\'e}rrez-Becker \and Christian Wachinger} %
\authorrunning{S.~P{\"{o}}lsterl et~al.} 
\institute{Artificial Intelligence in Medical Imaging (AI-Med),\\ Department of Child and Adolescent
Psychiatry,\\ Ludwig-Maximilians-Universit{\"{a}}t, Munich, Germany\\
\email{sebastian.poelsterl@med.uni-muenchen.de}} %
\maketitle              
\begin{abstract}
We introduce a wide and deep neural network for prediction of progression from
patients with mild cognitive impairment to Alzheimer's disease. Information from anatomical
shape and tabular clinical data (demographics, biomarkers) are fused in a single neural network. The network is
invariant to shape transformations and avoids the need to identify point correspondences between
shapes. To account for right censored time-to-event data, i.e., when it is only known that a patient
did not develop Alzheimer's disease up to a particular time point, we employ a loss commonly used in
survival analysis. Our network is trained end-to-end to
combine information from a patient's hippocampus shape and clinical biomarkers.
Our experiments on data from the Alzheimer's Disease Neuroimaging
Initiative demonstrate that our proposed model is able to learn a
shape descriptor that augments clinical biomarkers and outperforms
a deep neural network on shape alone
and a linear model on common clinical biomarkers.
\end{abstract}
\section{Introduction}
Alzheimer's disease (AD) is a neurodegenarative disorder
and the most common form of dementia diagnosed in people
over 65 years of age.
Initially, patients suffer from short memory loss, until progressive
deterioration eventually requires patients to be completely dependent
upon caregivers due to severe impairment of cognitive and motor
abilities~\cite{McKhann2011,Sperling2011,Albert2011}.
Mild cognitive impairment (MCI) is a pre-dementia stage which
is characterized by clinically significant cognitive decline, but
without impairing daily live~\cite{Petersen2011,Langa2014}.
Although subjects with MCI are at an increased risk of
developing dementia due to AD, a significant
portion of patients with MCI remain stable and do not progress~\cite{Petersen2011}.
The pathophysiological processes of this transition are
complex and not fully understood, but previous studies showed
that changes in certain biomarkers precede the onset
of cognitive symptoms by many years~\cite{Jack2013}.
Important biomarkers include brain atrophy measured by
magnetic resonance images (MRI),
levels of cortical amyloid deposition obtained
from cerebrospinal fluid (CSF),
and glucose uptake of neurons measured by
fluorodeoxyglucose positron emission tomography (FDG-PET)
(see \cite{Scheltens2016} for a detailed overview).
To stop or slow down the progression to dementia, it is vital
to identify those patients that are at an
increased risk for rapid progression from MCI to AD.
In particular, several previous studies have established strong morphological changes
in the hippocampus associated to the progression of dementia
\cite{Frisoni2008,Gerardin2009,wachinger2016whole,wachinger2016domain,gutierrez2018deep}.

We study progression to Alzheimer's disease by explicitly modelling the timing of this transition and by
considering the finite follow-up time and drop-out of patients
in clinical studies using techniques from survival analysis
(also called time-to-event analysis).
Survival analysis differs from traditional machine learning in the fact that
parts of the training data can only be partially observed -- they are \emph{censored}.
If a patient withdraws from the study, is lost to follow-up, or did
not develop AD during the study period, the patient's time of progression is \emph{right censored}, i.e., it is unknown whether
the patient has or has not progressed after the study ended.
Only if a patient develops AD during the study period, one can
record the exact time of this event -- it is \emph{uncensored}.

In this paper, we propose for the first time a wide and deep neural network
for survival analysis that learns to identify patients at high
risk of progressing to AD by fusing information from 3D hippocampus shape
and tabular clinical data.
To the best of our knowledge, no one has previously attempted to learn a deep
survival model on 3D anatomical shape representations in an end-to-end fashion.
In our experiments on data from the Alzheimer's Disease Neuroimaging Initiative,
we demonstrate by fusing information we can more accurately predict AD
converters than a baseline deep network on shapes and
a Cox's proportional hazards model on clinical data.
\section{Related Work}
Most previous work formulates progression analysis
from MCI to AD as a
classification problem within a fixed time horizon
such as 3 years (see e.g.~\cite{Beheshti2017,Chetelat2005,Cuingnet2011,Moradi2015,Tong2017}).
The major downside of this approach is that such a model cannot
generalize to other time spans, and that censored conversion
times are ignored during training.
Instead, it is statistically more appropriate to explicitly
incorporate censored event times using methods from survival analysis.
Several authors used survival analysis techniques
by combining information from various modalities such as
structural MRI, FDG-PET, genetics, and neuropsychological tests
\cite{Barnes2014,Da2014,Desikan2009,Desikan2010,Devanand2007,Kauppi2018,Li2018,Liu2017,Teipel2015,Vemuri2011,wachinger2016whole,Zhou2019}.
All of these approaches compute features from high-dimensional
imaging data in a pre-processing step, before training
a linear survival model.
They differ with respect to the type and extend of computed features,
which range from volume measurements of a few brain regions~\cite{Devanand2007}
to voxel-based analysis~\cite{Vemuri2011}.
In addition, we note that extensive prior work aims
to identify healthy controls, patients with MCI,
and patients with AD by casting it as
a three-way classification problem and using
multi-view machine learning techniques;
we refer interested readers to the review in~\cite{Liu2018}.

In contrast, this work focuses on multi-view learning to predict
progression from MCI to AD, which
has been formulated as a classification problem within a fixed
time period in~\cite{Liu2017a,Thung2016,Zhang2012,Zhou2019a}.
\cite{Zhang2012} propose to use sparsity-inducing penalties to combine
features extracted from MRI and PET images with CSF
measurements and neuropsychological tests.
MCI to AD conversion within 2 years was studied in~\cite{Liu2017a}.
They propose to learn from features extracted from MRI and FDG-PET,
and CSF measurements by view-aligned hypergraph learning.
The approach in~\cite{Thung2016} uses stability-weighted
low-rank matrix completion to impute missing values in
MRI and PET features, and neuropsychological tests.
They consider right censored conversion times as missing values
and try to impute the actual (unobserved) time of conversion
via matrix completion.
In~\cite{Zhou2019a}, the authors propose a missing-data-aware
approach to learn from MRI, PET, and genetics by learning a
common and multiple modality-specific latent feature representations.
To the best or our knowledge, the only previous work that employed
multi-view learning for survival analysis was presented in~\cite{Poelsterl2016}
for predicting adverse events in cancer and heart disease.

Using neural networks for survival analysis originated in the late 1990s
in the work of \cite{Bakker1999,Biganzoli1998,Faraggi1995,Liestol1994},
who studied relatively simple networks with one hidden layer applied
to tabular data.
The first deep survival model was proposed in~\cite{Katzman2018} and
builds on the loss proposed in~\cite{Faraggi1995}.
The only previous work that investigated deep learning for
MCI to AD conversion from multi-modal data is~\cite{Lee2019,Lu2018}.
Both approaches consider a classification problem within a
fixed time frame, which ignores censoring of conversion times.
In addition, the features in~\cite{Lee2019} were pre-computed from MRI
and not learned end-to-end.
In~\cite{Lu2018}, a deep network is proposed that learns
from 3D patches of MRI and FDG-PET at multiple scales.

Finally, \cite{gutierrez2018deep} proposed
a deep neural network operating on point clouds of multiple neuroanatomical shapes.
They study diagnosis of MCI and AD patients rather than progression,
and do not consider demographics or clinical biomarkers in their model.
\section{Methods}
We present a wide and deep neural network for learning from
right censored time-to-event data (see fig.~\ref{fig:pointnet}).
Our model takes a point cloud representation of an anatomical shape
and tabular data as input.
The deep part of the network is a PointNet~\cite{Qi_2017_CVPR}
that learns features describing the 3D geometric structure
of the left hippocampus.
The wide part of the network takes demographics and clinical biomarkers
and their interactions.
The network is trained to fuse both types of information in and
end-to-end fashion using a survival analysis loss appropriate for
right censored event times.
First, we are going to describe PointNet, which constitutes the
deep part of the network, before showing how it can be integrated
with tabular clinical data for survival analysis.
\subsection{Learning from Anatomical Shape}
\begin{figure}[tb]
\centering
\includegraphics[width=\textwidth]{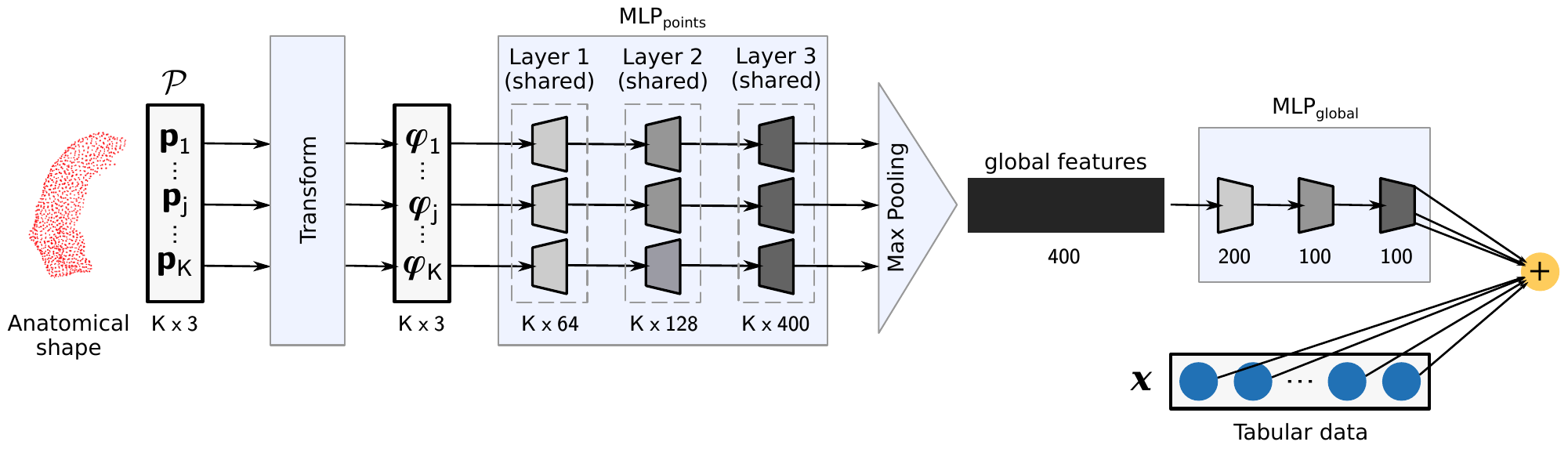}
\caption{\label{fig:pointnet}%
Wide and Deep PointNet Architecture.
The network	takes a point cloud representation $\set{P}$ of
the left hippocampus with $K$ points, applies a transformation,
and then aggregates point features by max pooling.
The global feature vector is processed by a global MLP outputting
a 100-dimensional latent representation that is fused with tabular
clinical data using a linear model.}
\end{figure}

We represent anatomical shapes as point clouds that represent
a 3D geometric structure as a set of coordinates.
Point clouds avoid the combinatorial irregularities and complexities
of meshes, and thus are easier to learn from.
However, the network needs to be constructed in a way to
consider that a point cloud is just an unordered set of points
that is invariant to permutations of its members.
To this end, we employ PointNet~\cite{Qi_2017_CVPR}, which
is illustrated in fig.~\ref{fig:pointnet} and described in more detail below.

The $i$-th point cloud $\set{P}_i$ is represented by a set of $K$ 3D coordinates
$\set{P}_i = \{\mathbf{p}_{i_1}, \ldots, \mathbf{p}_{i_K}\}$ with
$\mathbf{p}_{i_k} \in \bbbr^3$ being the $x$,
$y$, and $z$ coordinates.
To be invariant to permutations of the input set, the symmetric max
pooling operator across all embedding vectors of points is used.
We first pass each individual coordinate vector through a multilayer
perceptron $\mathrm{MLP}_\text{point}$ with shared weights among all points,
thus projecting each 3D point to a higher dimensional representation.
These representations are aggregated using the max pooling operator
across all points, which ensures that our downstream survival analysis
task is invariant to permutation:
\begin{equation}\label{eq:pointnet-vanilla}
\mathrm{POINTNET}(\set{P}_i) =
\mathrm{MAXPOOL}\left(
\mathrm{MLP}_\text{points}(\mathbf{p}_{i_1}),
\ldots,
\mathrm{MLP}_\text{points}(\mathbf{p}_{i_K})
\right) .
\end{equation}
$\mathrm{MLP}_\text{point}$ is a three-layer network
with 64, 128, and 400 dimensional outputs, respectively,
with rectified linear units (ReLU) and batch normalization
\cite{Ioffe2015}.
Hence, we extract 400 features that globally describe
the input anatomical shape.

In order to make our network invariant to rotation of the input point cloud,
we use an affine transformation network that outputs a
rotation matrix $\mathbf{T} \in \bbbr^{3 \times 3}$
which is multiplied by the raw 3D coordinates of input points.
This transformation is learned in a data-dependent manner
by using an additional $\mathrm{POINTNET}$ network that
learns to predict the optimal $\mathbf{T}$ for each
individual point cloud.
The global feature vector computed by $\mathrm{POINTNET}$ is fed
to three fully-connected layers with 200, 100, and 9 units,
ReLU activation function and batch normalization, respectively.
Finally, we modify the vanilla PointNet in \eqref{eq:deep-wide-net}
by transforming individual points by the output of the
transformation network:
\begin{equation}\label{eq:pointnet}
\begin{split}
\mathrm{TRANSFORM}(\set{P}_i) &=
\mathrm{MAXPOOL}\left(
\mathrm{MLP}_\text{points}(\mathbf{p}_{i_1}),
\ldots,
\mathrm{MLP}_\text{points}(\mathbf{p}_{i_K})
\right) ,\\
\bm{\varphi}_{i_k} &= \mathrm{TRANSFORM}(\set{P}_i)\mathbf{p}_{i_k} ,\\
\mathrm{POINTNET}(\set{P}_i) &=
\mathrm{MAXPOOL}\left(
\mathrm{MLP}_\text{points}(\bm{\varphi}_{i_1}),
\ldots,
\mathrm{MLP}_\text{points}(\bm{\varphi}_{i_K})
\right) .
\end{split}
\end{equation}

\subsection{Wide and Deep Neural Network}
After obtaining a global latent representation of an anatomical shape,
we can further learn high-level
descriptors of point clouds by feeding the output of the
max pooling operation to an MLP.
In addition, we can leverage routine clinical patient information
to predict progression to Alzheimer's disease.
Typically, such information consists of feature vectors
that are either dense (e.g. biomarker concentrations), or sparse
(e.g. one-hot encoded genetic alterations).
Compared to individual points in a point cloud, clinical
information already contains rich information for which we do not
need to learn a highly abstract latent representation.
In fact, most clinical
research relies on linear models, which allow for easy interpretation
of individual feature's contribution to the overall prediction.

Here, we jointly train a linear
model on clinical information with a deep PointNet on
anatomical shapes using a wide and deep architecture~\cite{Cheng2016}.
While the deep component learns a complex latent representation
of anatomical shape, the linear component models
known clinical variables $\mathbf{x} \in \bbbr^d$ associated with
Alzheimer's disease.
In particular, we can easily incorporate
gene-gene (epistasis) and gene–environment interactions by using
a cross-product transformation $\phi(\mathbf{x})$~\cite{Cheng2016}.
Thus, the final patient-level latent representation is given by
\begin{multline}\label{eq:deep-wide-net}
\mu(\mathbf{x}_i, \set{P}_i) = \mathbf{w}_\text{wide}^\top
\mathrm{CONCAT}
\left( \mathbf{x}_i, \phi(\mathbf{x}_i) \right) \\
+ \mathbf{w}_\text{deep}^\top
\mathrm{MLP}_\text{global}(
\mathrm{POINTNET}(\set{P}_i)) ,
\end{multline}
where $\mathrm{CONCAT}$ denotes vector concatenation,
$\mathrm{POINTNET}$ is the global feature vector from \eqref{eq:pointnet},
$\mathrm{MLP}_\text{global}$ is a three-layer MLP with
200, 100, and 100 units, ReLU activation and batch normalization, and
$\mathbf{w}_\text{wide}$ and $\mathbf{w}_\text{deep}$
are weights to be learned.
\subsection{Survival Analysis}
Our overall objective is to predict progression from
mild cognitive impairment to Alzheimer's disease from
right censored time-to-event data, which demands
for proper training algorithms that take this unique characteristic
into account.
More formally, we denote by
$t_i > 0$ the time of an event (Alzheimer's disease),
and $c_i > 0$ the time of censoring of the $i$-th patient.
Due to right censoring, it is only possible to observe
$y_i = \min(t_i, c_i)$ and $\delta_i = I(t_i \leq c_i)$ for every patient,
with $I(\cdot)$ being the indicator function and $c_i = \infty$ for
uncensored records.
Hence, training our survival model is based on a dataset comprising
quadruplets $(\set{P}_i, \mathbf{x}_i, y_i, \delta_i)$
for $i = 1,\ldots,n$.
After training, the survival model ought to predict a risk score of
experiencing an event based on a point cloud and a set of
clinical features.
As loss function, we employ the loss proposed in~\cite{Faraggi1995},
which is an extension of Cox's proportional hazards model~\cite{Cox1972}
to neural networks.
Let $\bm{\Theta}$ denote the set of all parameters of
the wide and deep neural network~\eqref{eq:deep-wide-net},
then we want to solve
\begin{equation}
\argmin_{\bm{\Theta}}\quad \sum_{i=1}^n \delta_i \left[
\mu(\vec{x}_i, \set{P}_i \,|\,\bm{\Theta})
- \log \left( \sum_{j \in \set{R}_i} \exp(
\mu(\vec{x}_j, \set{P}_i\,|\,\bm{\Theta}) )
\right) \right],
\end{equation}
where $\set{R}_i = \{ j\,|\,y_j \geq t_i \}$ denotes the risk set,
i.e., the set of patients who were still free of Alzheimer's disease
shortly before time point $t_i$.
\section{Experiments}
\subsection{Data}
In our experiments, we are using data from the Alzheimer's Disease Neuroimaging
Initiative (ADNI)~\cite{Jack2008}.
ADNI was launched in 2003 as a public-private partnership
with the primary goal to test whether longitudinal MRI and PET imaging
combined with other biomarkers, clinical and neuropsychological assessments
to measure the progression of MCI and early AD.
For up-to-date information, see \url{www.adni-info.org}.
We selected 397 subjects with MCI at baseline and at least one
follow-up visit.
Magnetic resonance images of all subjects
were processed with FreeSurfer~\cite{Fischl2012} to obtain segmentations,
which were subsequently pre-processed using the grooming operations included in
ShapeWorks~\cite{Cates2008} to obtain smooth hippocampi surfaces.
We used left hippocampus shapes represented as point clouds comprised of 1024 points.
For tabular clinical data, we used age, gender, education,
CSF, FDG-PET, and AV45-PET.
CSF measurements included levels of beta amyloid 42 peptides (A$\beta_{42}$),
total tau protein (T-tau), and
Tau phosphorylated at threonine 181 ($\text{p-Tau}_{181}$).
We augment age to account for non-linear effects by using
a natural B-spline expansion with four degrees of freedom
and an interaction term between age and gender~\cite{Hastie2009}.
Education, which is a categorical variable, was encoded
using orthogonal polynomial coding.
In addition, we considered left hippocampus volume (normalized by intra-cranial volume)
as estimated by FreeSurfer~\cite{Fischl2012}
from MRI scans of the brain.
\subsection{Model Training}
We trained our deep and wide network using Adam~\cite{Kingma2014} for
120 epochs with weight decay.
We tuned hyper-parameters (size of PointNet's global feature vector,
size of $\mathbf{w}_\text{deep}$, weight decay, learning rate schedule,
$\beta_1$ of Adam) using Bayesian black-box optimization
by computing the model's performance on the validation set~\cite{liaw2018tune}.
Data is randomly split into three parts: 80\% for training,
10\% for validation, and 10\% for testing.
We repeated this process 10 times with different splits.
The performance of all methods was estimated by Harrell's
concordance index (c index), which is identical to the area
under the receiver operating characteristics curve if the
outcome is binary and no censoring is present~\cite{Harrell1982}.
As baseline model, we selected a linear Cox's proportional hazards
model~(CoxPH)~\cite{Cox1972} trained on tabular clinical data.
The baseline model was trained once on tabular clinical data only (see above),
and once with the volume of left hippocampus
included as additional feature.
We note that CoxPH and our model optimize the same loss during training.
Therefore, differences in performance stem from the ability of our model
to directly incorporate 3D anatomical shape information.
\section{Results}
\begin{figure}
\centering
\includegraphics[scale=.4]{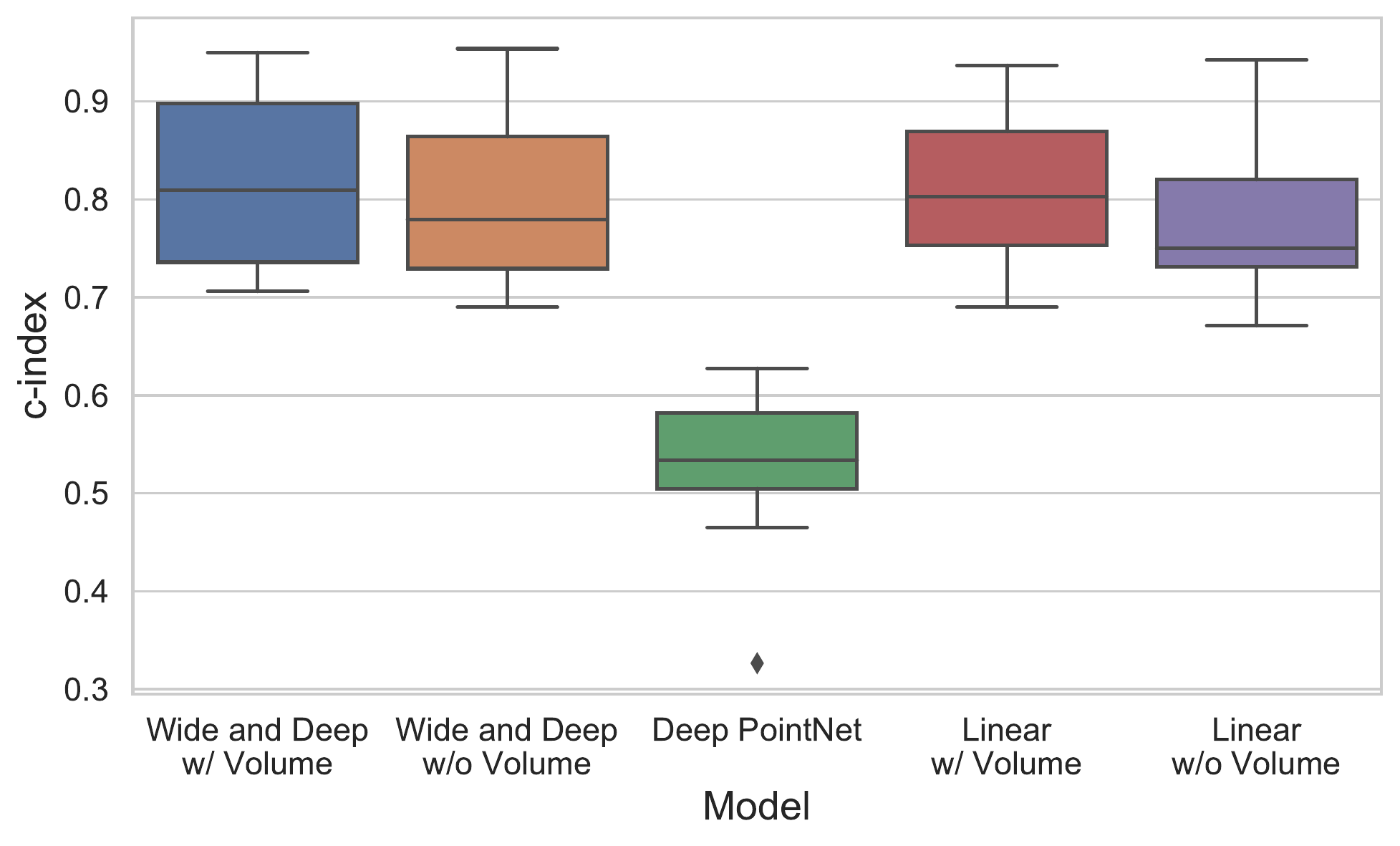}
\caption{\label{fig:results}%
Performance of individual models across ten random splits of the data.
w/ Volume: tabular data includes left hippocampus volume.
w/o Volume: tabular data does \emph{not} include left hippocampus volume.}
\end{figure}
The performance of our deep and wide network and baseline models is
summarized in fig.~\ref{fig:results}.
It shows that tabular clinical makers with a median $c$ index of 0.750
are already strong predictors of conversion from MCI to AD.
When including hippocampus volume as additional feature, the
median $c$ index increased to 0.803.
Using a deep PointNet solely using hippocampus shape and ignoring
any clinical variables resulted in a $c$ index of 0.534.
Our deep and wide network achieved a median $c$ index of 0.780
without hippocampus volume, and 0.809 with hippocampus volume.
The latter is the model with highest median $c$ index and
outperforms the linear model with hippocampus volume on 6 of 10
splits.
This shows that when jointly learning a deep PointNet, it is able to learn a powerful global
descriptor of hippocampus shape that augments clinical features
for MCI-to-AD progression.
Moreover, our results confirm that hippocampus volume is a useful independent predictor
that cannot be fully captured by anatomical shape alone,
as described previously~\cite{wachinger2016domain}.
\begin{figure}[tb]
\centering
\includegraphics[scale=.4]{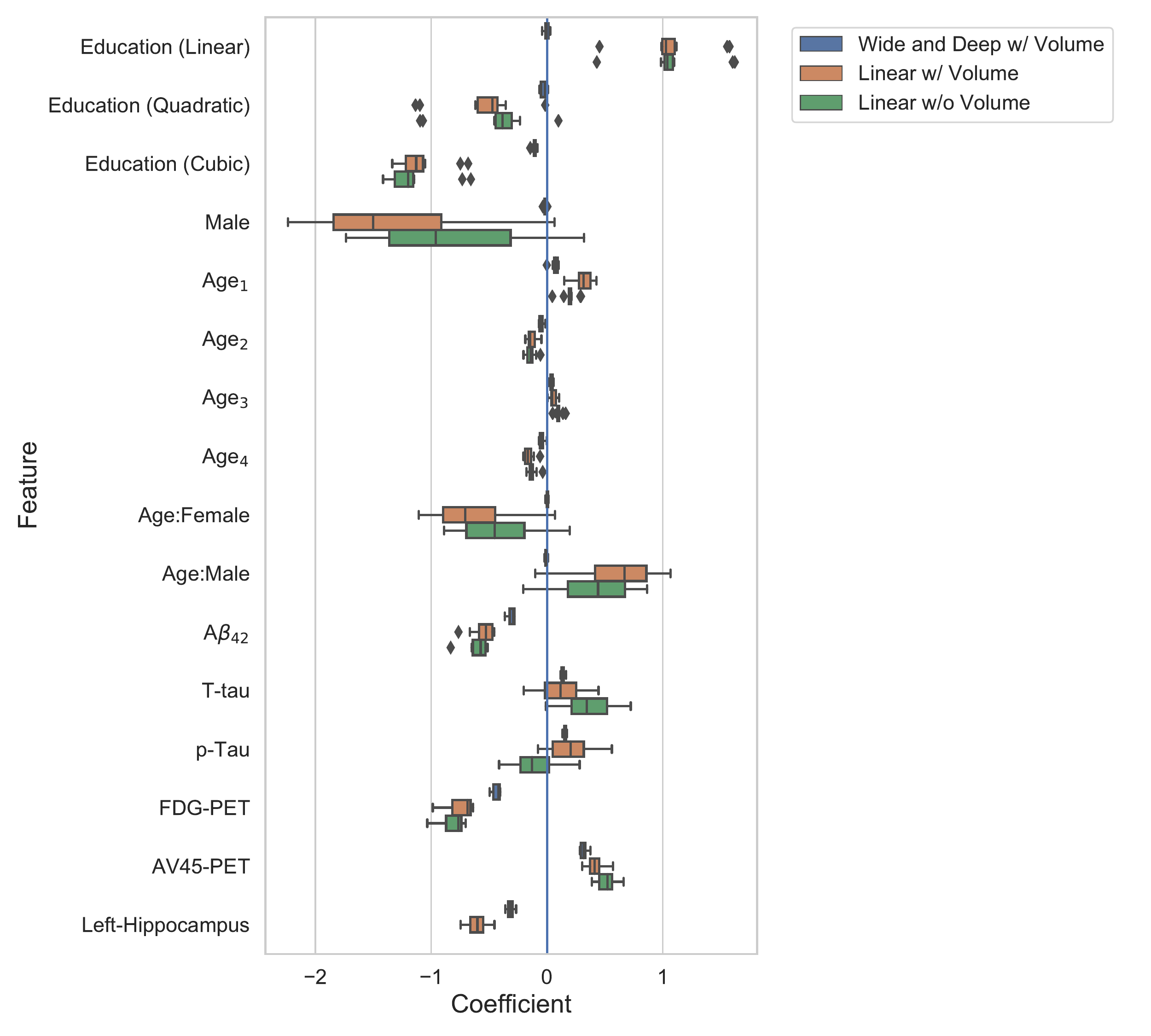}
\caption{\label{fig:coefficients}%
Comparison of coefficients associated with tabular clinical features.
Additional eight orthogonal polynomial encodings of education have been omitted from this plot.
w/ Volume: tabular data includes left hippocampus volume.
w/o Volume: tabular data does \emph{not} include left hippocampus volume.}
\end{figure}

We can also compare the coefficients of the linear models with the
linear part of our wide and deep neural network.
The coefficients can be directly interpreted in terms
of log-hazard ratio, which is a measure of effect a variable has
on survival, similar to log-odds ratio in logistic regression.
The coefficients across all folds are depicted in fig.~\ref{fig:coefficients}.
All models agree with respect to which features are contributing to increased/decreased
hazard of AD, as indicated by the coefficients' sign, except for p-Tau.
The linear model without hippocampus volume associated higher p-Tau levels with
a decrease in hazard (on average) compared to the other models,
which is surprising because hyperphosphorylation of tau is a marker for AD~\cite{Blennow2001}.
The most important clinical features (in terms of magnitude)
are gender and education for both linear models, but
have only minor importance for the deep and wide network.
Similar behavior can be observed for age-gender interactions.
In addition, increased hippocampus volume has a relatively high
importance and is associated with a decreased hazard of AD.
It is ranked third for the deep and wide network and eleven
for the linear model.
FDG-PET has the biggest effect for the wide and deep network
and is also among the top 4 features for the linear models.
From a clinical perspective, this result is reassuring as
reduction of metabolic activity in cortical regions has been
associated with AD~\cite{Minoshima1997}.
Finally, we note that the variability of coefficients across
splits is smaller for the deep and wide neural network compared
to the linear model. We believe this is an effect of using weight decay
during optimization, which penalizes large coefficients.
\section{Conclusion}
We proposed a wide and deep neural network that fuses 3D anatomical shape
and tabular clinical variables for the prediction of MCI-to-AD conversion.
We trained a model end-to-end using a survival loss that properly accounts
for right censored time of conversion.
Our experiments demonstrate that the proposed architecture is able
to learn a global shape descriptor that augments clinical variables
and leads to improved prediction performance.
\subsubsection*{Acknowledgements}
This research was partially supported by the Bavarian State Ministry of Education,
Science and the Arts in the framework of the Centre Digitisation.Bavaria (ZD.B).
We gratefully acknowledge the support of NVIDIA Corporation with the donation of
the Quadro P6000 GPU used for this research.
\bibliographystyle{splncs03}
\bibliography{references.bib}
\end{document}